\documentclass[letterpaper, 10 pt, conference]{ieeeconf}  

\IEEEoverridecommandlockouts                              

\usepackage{graphics} 
\usepackage{epsfig} 
\usepackage{mathptmx} 
\usepackage{times} 
\usepackage{amsmath} 
\usepackage{amssymb}  
\usepackage{color}
\usepackage{todonotes}
\usepackage{algorithm}
\usepackage{algpseudocode}
\usepackage{caption}
\usepackage{afterpage}
\usepackage{placeins}
\usepackage{booktabs}
\usepackage{siunitx}
\usepackage{float}
\usepackage{flushend}
\usepackage{algorithmicx}
\usepackage{algorithm}
\usepackage{algpseudocode}
\usepackage{hyperref}

\newcommand{\etal}{\textit{et al}.}

\algnewcommand\algorithmicinput{\textbf{Input:}}
\algnewcommand\Input{\item[\algorithmicinput]}
\algnewcommand\algorithmicoutput{\textbf{Output:}}
\algnewcommand\Output{\item[\algorithmicoutput]}

\newcommand{\singleQuote}[1]{\lq{#1}\rq}

\graphicspath{{figures/}}

\title{\LARGE \bf
Backtracking Regression Forests for Accurate Camera Relocalization
}

\author{Lili Meng$^{1}$, Jianhui Chen$^{2}$, Frederick Tung$^{2}$,  James J. Little$^{2}$,  Julien Valentin$^{3}$, Clarence W. de Silva$^{1}$ 
\thanks{$^{1}$ Lili Meng and Clarence W. de Silva are with the Department of Mechanical Engineering, The University of British Columbia, Vancouver, Canada.
        {\tt\small \{lilimeng,desilva\}@mech.ubc.ca}  }%
\thanks{ $^{2}$ Jianhui Chen, Frederick Tung and James J. Little are with the Department of Computer Science, The University of British Columbia, Vancouver, Canada.
        {\tt\small \{jhchen14,ftung,little\}@cs.ubc.ca} }%
\thanks{$^{3}$ Julien Valentin is with Perceptive$^{IO}$ Inc. United States.  {\tt\small  julien@perceptiveio.com} }
\thanks{The authors would like to thank Jamie Shotton, Torsten Sattler and Alex Kendall for helpful discussions.}
} 

\begin{document}

\maketitle

\thispagestyle{empty}
\pagestyle{empty}

\begin{abstract}
Camera relocalization plays a vital role in many robotics and computer vision tasks, such as global localization, recovery from tracking failure, and loop closure detection. Recent random forests based methods directly predict 3D world locations for 2D image locations to guide the camera pose optimization.  During training, each tree greedily splits the samples to minimize the spatial variance. However, these greedy splits often produce uneven sub-trees in training or incorrect 2D-3D correspondences in testing. To address these problems, we propose a sample-balanced objective to encourage equal numbers of samples in the left and right sub-trees, and a novel backtracking scheme to remedy the incorrect 2D-3D correspondence 
predictions. Furthermore, we extend the regression forests based methods to use local features in both training and testing stages for outdoor RGB-only applications. Experimental results on publicly available indoor and outdoor datasets demonstrate the efficacy of our approach, which shows superior or on-par accuracy with several state-of-the-art methods.
\end{abstract}


\section{INTRODUCTION}
Camera relocalization plays a critical role in robotics and augmented reality (AR), especially in indoor environments where GPS is not available. Light-weight and affordable visual sensors make camera relocalization algorithms an appealing alternative to GPS or other expensive sensors. Recent consumer robotics products such as iRobot 980 and Dyson 360 eyes are already equipped with visual simultaneous localization and mapping (SLAM) techniques, enabling them to effectively navigate in complex environments. In order to insert virtual objects in an image sequence (i.e.,\ in AR applications), camera poses have to be estimated in a visually acceptable way. 


Random forests based methods are among the first machine learning methods for camera relocalization \cite{shotton2013scene, guzman2014multi,valentin2015exploiting}. In these methods, the forest is trained to directly predict correspondences from image pixels to points in the scene's 3D world coordinate. These correspondences are then used for camera pose estimation based on an efficient RANSAC algorithm without an explicit 3D model of the scene, which is very attractive when a 3D model is not available.  The latest work \cite{Brachmann_2016_CVPR, LiliRandom} extends these random forests based methods to test with just RGB images. However, depth images are still necessary to get 3D world coordinate labels in the training stage. 
In the training stage of these random forests based methods, each tree greedily splits the samples to minimize the spatial variance. However, these greedy splits usually produce uneven left and right sub-trees, or incorrect 2D-3D correspondences in the test.  To address this problem, we propose a sample-balanced objective to encourage equal numbers of samples in the left and right sub-trees, and a novel backtracking scheme to remedy the incorrect 2D-3D correspondences caused by greedy splitting. We demonstrate the efficacy of our methods through evaluations on publicly available indoor and outdoor datasets.

\begin{figure}
	\begin{center}
		\includegraphics[width = 1\linewidth]{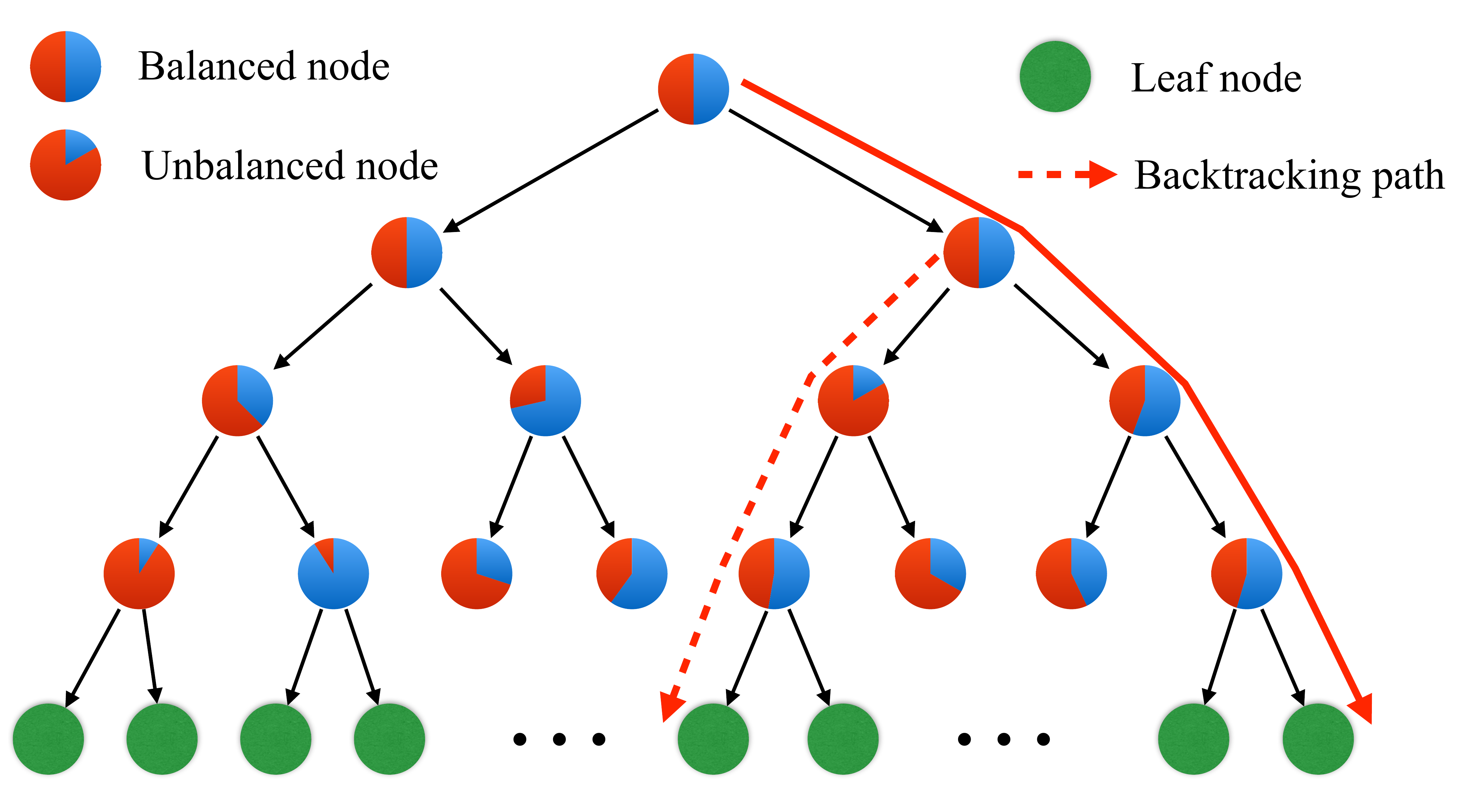}  
	\end{center}
	\vspace{-4mm}
	\caption{\textbf{Decision tree using two split objectives and a backtracking method.} The split nodes are illustrated as the pie charts which show the percentage of samples in the left and right sub-trees. In this five-level tree, the first two levels are split using the sample-balanced objective, while the rest of the levels are split using the unbalanced objectives (i.e., the spatial-variance objective). In testing, our method backtracks the tree to find the optimal prediction.} 
	\label{fig:tree_example}
    \vspace{-4mm}
\end{figure}

Some recent methods based on deep learning \cite{kendall2015posenet,kendall2016modelling,walch2016image} overcome the challenges of using depth images for camera relocalization, extending their application to outdoor scenes. These methods train a convolutional neural network to regress the 6-DOF camera pose from a single RGB image in an end-to-end manner in real time. However, even integrating LSTM units on the CNN output \cite{walch2016image}, these methods still lead to much lower accuracy compared with methods based on random forests \cite{shotton2013scene} in indoor scenes and local-feature based methods \cite{sattler2016efficient} in outdoor scenes in general. To eliminate the dependency on depth images while ensuring high accuracy, we integrate local features in the random forests, broadening the application of random forests methods to outdoor scenes for the first time while achieving the best accuracy against several strong state-of-the-art baselines. 

To summarize, our main contributions in this work are as follows:
\begin{itemize} 
\item We propose a sample-balanced objective that encourages equal numbers of samples in the sub-trees, increasing prediction accuracy while reducing training time.
\item We propose a novel backtracking scheme to remedy incorrect 2D-3D correspondence predictions caused by greedy splitting, further improving prediction accuracy.
\item We integrate local features in the regression forests, enabling the use of RGB-only images for both training and testing. The elimination of the dependence on depth images broadens the scope of application from indoor to outdoor scenes.  
\end{itemize}
\section{Related Work}
Camera relocalization is one of the fundamental computer vision and robotics problems. It has been widely studied in the context of large scale global localization \cite{nister2006scalable, torii201524, kendall2015posenet}, recovery from tracking failure \cite{klein2007parallel,glocker2015real}, loop closure detection in visual SLAM \cite{whelan2015elasticfusion}, global localization in mobile robotics \cite{se2005vision, cummins2011appearance}, and sports camera calibration \cite{chen2017where,LiliRandom, chen2015mimicking}. 
Methods based on local features, keyframes, random forests and deep learning are four general categories of camera relocalization,  although other successful variants and hybrid methods \cite{rubio2015efficient} exist.

{\bf Local feature based approaches}
  Local feature based methods  \cite{se2005vision, nister2006scalable} usually match the descriptors extracted from the incoming frame and the descriptors from frames in the database. The $3$D locations of the frames in world coordinates are also stored together with their descriptors. Then the combination of the perspective-three-point \cite{gao2003complete} and RANSAC \cite{fischler1981random} is usually employed to determine the camera pose. The local features, such as commonly used SIFT \cite{lowe2004distinctive}, can represent the image local properties, so they are more robust to viewpoint changes as long as a sufficient number of keypoints can be recognized. However, these methods require a large database of descriptors and efficient retrieval methods.  Much recent work has focused on efficiency \cite{sattler2016efficient}, scalability \cite{schindler2007city}, and learning of feature detection, description, and matching \cite{lepetit2006keypoint}.

\textbf{Keyframe based approaches}
Methods based on keyframes \cite{gee20126d,engel2014lsd} hypothesize an approximate camera pose by computing whole-image similarity between a query image and keyframes. For example, randomized ferns encode an RGB-D image as a string of binary codes, which have been used to recover from tracking failure \cite{glocker2015real} and loop closure in SLAM \cite{whelan2015elasticfusion}. However, these keyframe-based methods provide inaccurate matches when the query frame is significantly different from the stored keyframes. 

{\bf Random forests based approaches}
Approaches based on random forests have gained interest since the introduction of scene coordinate regression forests (SCRF) \cite{shotton2013scene}. Random forests are used as regression for camera relocalization so we also refer to them as regression forests. These approaches first employ a regression forest to learn an estimate of each 2D image pixel's corresponding 3D point in the scene's world coordinates. Then the camera pose optimization is conducted by an adapted version of preemptive RANSAC. A hybrid discriminative-generative learning architecture is proposed in \cite{guzman2014multi} to overcome the many local optima problem of non-convex optimization in SCRF. A method to train and exploit uncertainty from regression forests for accurate camera relocalization is presented in  \cite{valentin2015exploiting}.  \cite{LiliRandom,valentin2016learning,Brachmann_2016_CVPR} extended the random forest based methods to use RGB images at test time, while depth images are still needed to get ground truth labels. \cite{valentin2016learning} proposed multiple accurate initial solutions and defined a navigational structure over the solution space that can be used for efficient gradient-free local search. However, it needs an explicit 3D model as the auxiliary of the RGB image at test time. Only RGB pixel comparison features are used in the forest in \cite{LiliRandom}, eliminating the dependency on the depth image at test time. Furthermore, it integrates random RGB features and sparse feature matching in an efficient and accurate way, allowing the method to be applied for fast sports camera calibration in highly dynamic scenes. SCRF has been extended \cite{cavallari2017fly} for online camera relocalization by adapting a pre-trained forest to a new scene on the fly. 

{\bf Deep learning based approaches}
Deep learning has led to rapid progress and impressive performance on a variety of computer vision tasks, such as visual recognition and object detection \cite{redmon2016you}. There are emerging works \cite{kendall2015posenet,kendall2016modelling,walch2016image} for the camera relocalization task.
PoseNet \cite{kendall2015posenet} trains a convolutional neural network to regress the 6-DOF camera pose from a single RGB image in an end-to-end manner in real time. Bayesian PoseNet \cite{kendall2016modelling} applies an uncertainty framework to the PoseNet by averaging
Monte Carlo dropout samples from the posterior Bernoulli  distribution  of  the  Bayesian ConvNet's weights. Besides the 6-D pose, it also estimates the model's relocalization uncertainty. To improve the PoseNet accuracy, geometric loss function is explored in \cite{kendall2017geometric} and LSTM units are used on the CNN output to capture the contextual information. However, these deep learning based methods have much lower overall accuracy compared with random forests based methods in indoor scenes and local feature based methods in outdoor scenes. Moreover, high-end GPU dependency and high power consumption make these deep learning based methods less favorable for energy sensitive applications such as mobile robots.

\section{Method}
\label{sec:method}

We model the world coordinate prediction problem as a regression problem:
\begin{equation}
\hat{\mathbf{m}}_{\mathbf{p}} = f_{r} (\mathtt{I}, \mathtt{D}, \mathbf{p} | \mathbf{\theta})
\end{equation}
\normalsize
where $\mathtt{I}$ is an RGB image, $\mathtt{D}$ is an optional depth image, $\mathbf{p} \in \mathbb{R}^2$ is a 2D pixel location in the image $\mathtt{I}$, $\mathbf{m}_{\mathbf{p}} \in \mathbb{R}^3$ is the corresponding 3D scene coordinate of $\mathbf{p}$, and $\mathbf{\theta}$ is the model parameter. In training, $\{\mathbf{p}, \mathbf{m}_\mathbf{p} \}$ are paired training data. In testing, the 3D world location $\hat{\mathbf{m}}_{\mathbf{p}}$ is predicted by the model $\mathbf{\theta}$. Then, the camera parameters are estimated using predicted image-world correspondences in a RANSAC framework.

The novelty of our method lies in both the regression forest training and prediction. First, besides the spatial-variance objective, we add a sample-balanced split objective in the training stage. This objective function encourages equal numbers of samples in the left and right sub-trees.  It decreases the correlation between individual trees so that the generalization error of the whole forest drops \cite{breiman2001random}. In practice, it acts as a regularization term in the training and thus improving the prediction accuracy. Second, we present a novel backtracking technique in prediction. The backtracking searches the tree again using a priority queue and decreases the chance of falling in a local minimum.

\subsection{Weak Learner Model}

A regression forest is an ensemble of decision trees $T$ independently trained to minimize a regression error (e.g.,\ 3D location error). Each tree is a binary tree consisting of split nodes and leaf nodes. Each split node $i$ represents a weak learner parameterized by $\theta_i=\{\mathbf{\phi}_i, \tau_i\}$ where $\mathbf{\phi}_i$ is one dimension in features and $\tau_i$ is a threshold. 

In training, the decision tree optimizes the parameters $\theta_i$ of each node from the root node and recursively processes the child nodes until the leaf nodes by optimizing a chosen objective function as follows:

\begin{equation}
\label{eq:binary}
h(\mathbf{p};\theta_i)=
\left\{
\begin{aligned}
0, & \quad \text{if} \ f_{\phi_i}(\mathbf{p}) \leq \tau_i, \quad \text{go to the left subset $S_i^L$}\\
1, & \quad \text{if} \ f_{\phi_i}(\mathbf{p}) > \tau_i, \quad \text{go to the right subset $S_i^R$}
\end{aligned}
\right.
\end{equation}

$f_{\phi_i}(\mathbf{p})$ represents a general feature that is associated with pixel location $\mathbf{p}$, the weak learner model can use various features according to  application scenarios. Table \ref{table:features} shows image features that are used in our method. More details on the image features are provided in Sec. \ref{sec:evaluations} where more context information is available.

\begin{table}
\begin{center}
\scalebox{1}{
\begin{tabular}{|l|ccc|}
\hline
      &\multicolumn{3}{c}{\textbf{Feature Types}}   \vline \\
    Dataset      & Split & Backtrack & Local descriptor \\
\hline
    Indoor RGB-D       & Random & Random   & WHT \\ 
    Outdoor RGB      & SIFT   & SIFT     & SIFT\\     
\hline
\end{tabular}
}
\end{center}
\vspace{-1mm}
  \caption{\textbf{Image features}. We use random features \cite{shotton2013scene}, SIFT features \cite{lowe2004distinctive} and Walsh-Hadamard transform (WHT)  features \cite{hel2005real} for indoor and outdoor camera relocalization.}
   \label{table:features}
\vspace{-3mm}
\end{table}

\subsection{Training Objectives}
\label{subsec:objective}
Our regression forest uses two objectives. At upper levels of a tree, we use a sample-balanced objective:
\begin{equation}
Q_b(S_n, \theta) = \frac{abs(|S_n^L| - |S_n^R|)}{|S_n^L| + |S_n^R|}
\label{equ:balance}
\end{equation}
where $abs(.)$ represents the absolute value operator and $|S|$ represents the size of set $S$. This objective penalizes uneven splitting. The sample-balanced objective has two advantages. First, it is faster as it only counts the number of samples in sub-trees. Second, it produces more accurate predictions in practice which is shown in experiments in Sec. \ref{sec:evaluations}. 

When the tree depth is larger than a threshold $L_{max}$, we use the spatial-variance objective:
\begin{equation}
\label{equ:redu_loss}
\begin{aligned}
Q_v(S_n, \theta) &= \sum_{d \in \{L,R \}} \frac{|S_n^d(\theta)|}{|S_n|} V(S_n^d(\theta)) \\
\mathrm{with} \;\; V(S) & = \frac{1}{|S|}\sum_{m \in S } \| m- \bar{m} \|_2^2
\end{aligned}
\end{equation}
where $\bar{m}$ is the mean of $m$ in $S$ and subset $S_n^d$ is conditioned on the split parameter $\theta$. The value $L_{max}$ is experimentally set.

At the end of training, our method stores the weak learner parameters $\theta_i$ in each split node. In a leaf node, our method not only stores a mean vector of 3D positions but also a mean vector of local patch descriptors. The local patch descriptor will be used to choose the optimal predictions in a backtracking process described in Sec.\ref{subsec:backtracking}.

\subsection{Backtracking in Regression Forests Prediction}
\label{subsec:backtracking}

\begin{algorithm}[t]
\caption{Decision Tree Prediction with Backtracking}
\label{alg:one_tree_prediction}
\begin{algorithmic}[1]
\Input A decision tree $T$, testing sample $S$, maximum number of leaf nodes to examine $N_{max}$
\Output Predicted label and minimum feature distance  
\Procedure{TreePrediction} {$T, S, N_{max}$}
\State $count \gets 0$
\State $PQ \gets $ empty priority queue
\State $R \gets \emptyset$ \Comment{the prediction result}

\State call \textsc{TraverseTree}$(S, T, PQ, R)$
\While{PQ not empty and $count < N_{max}$}
\State $T \gets$ top of $PQ$
\State call \textsc{TraverseTree}$(S, T, PQ, R)$
\EndWhile

\State \Return $R$
\EndProcedure

\Procedure{TraverseTree}{$S, T, PQ, R$}
\If{$T$ is a leaf node}
\State $dist \gets distance(T.feature, S.feature)$ \Comment{e.g.,\ L2 norm} 
\If {$R$ is $\emptyset$ or $dist < R.dist$}
\State $\{R.label, R.dist\} \gets \{T.label, dist\}$ \Comment{update}
\EndIf
\State $count = count + 1$

\Else
\State $splitDim \gets T.splitDim$
\State $splitValue \gets T.splitValue$
\If{$S(splitDim) < splitValue$}
\State $closestNode \gets T.left$
\State $otherNode \gets T.right$
\Else
\State $closestNode \gets T.right$
\State $otherNode \gets T.left$
\EndIf

\State add $otherNode$ to $PQ$
\State call \textsc{TraverseTree}$(S, closestNode, PQ, R)$
\EndIf
\EndProcedure
\end{algorithmic}
\end{algorithm}

In the test phase, a regression tree greedily predicts 3D locations by comparing the feature value and the split values in the split nodes. Because the comparison is conducted on a single dimension at a time, making mistakes is inevitable. 

To mitigate this drawback, we propose a backtracking scheme to find the optimal prediction within time budget using a priority queue. The priority queue stores the sibling nodes that are not visited along the path when a testing sample travels from the root node to the leaf node. The priority queue is ordered by increasing distance to the split value of each internal node. The backtracking continues until a predefined number $N_{max}$ of leaf nodes are visited. Algorithm \ref{alg:one_tree_prediction} illustrates the detailed procedure of our backtracking technique. 

Fig. \ref{fig:tree_example} illustrates the idea of using two objectives and the backtracking in a five-level decision tree. The first two levels use the sample-balanced objective while the next two levels use the spatial-variance objective. In the testing, the backtacking checks multiple leaf nodes to find the optimal prediction.

\subsection{Camera Pose Optimization}
When depth images are available (i.e., indoor application), we use the Kabsch algorithm \cite{kabsch1976solution} to estimate camera poses. Otherwise, we use the \textit{solvePnPRansac} method from OpenCV. Because predictions from the regression forests may still have large errors, we use the preemptive RANSAC method \cite{nister2005preemptive}  to remove outliers.

\section{Evaluations}
\label{sec:evaluations}

In this section, we evaluate our method on publicly available indoor and outdoor datasets against several strong state-of-the-art methods. 

\subsection{Indoor Camera Relocalization}
\label{subsec:Indoor}
\vspace{-1mm}
\begin{table*}
\begin{center}
\scalebox{0.95}{
\begin{tabular}{|l|ll|l|cccc|cccc|}
\hline
         &\multicolumn{2}{c}{\textbf{Frame numbers}} \vline  & \textbf{Spatial}& \multicolumn{4}{c}{\textbf{Baselines}}\vline & \multicolumn{4}{c}{\textbf{Our Results}}  \vline \\ 
  \textbf{Sequence}&training& test & \textbf{Extent} & ORB+PnP & SIFT+PnP &Random+SIFT& MNG& BTBRF& UBRF & BRF &BTBRF \\
\hline
Training &--- & ---& & RGB-D & RGB-D & RGB-D&RGB-D &RGB-D & RGB-D &RGB-D&RGB-D\\ 
Test  &--- &---& & RGB  & RGB &RGB& RGB+3D model & RGB &RGB-D &RGB-D&RGB-D\\    
\hline\hline
Kitchen &744 &357&$33m^3$ & 66.39\% & 71.43\% &70.3\% & 85.7\%&77.1\% &82.6\% &88.2\% &\textbf{92.7}\%\\ 
Living  &1035 &493 & $30m^3$ & 41.99\% & 56.19\% & 60.0\%&71.6\%&69.6\% &81.7\% &90.5\% &\textbf{95.1}\%\\ 
\hline

Bed     &868 &244 & $14m^3$ & 71.72\% & 72.95\% & 65.7\% &66.4\%&60.8\% &71.6\% &81.3\%  &\textbf{82.8}\%\\ 
Kitchen &768 &230 &$21m^3$ & 63.91\% & 71.74\% & 76.7\% &76.7\% &82.9\% &80.0\% & 85.7\%&\textbf{86.2}\%  \\  
Living &725 &359 & $42m^3$ & 45.40\% & 56.19\% & 52.2\%&66.6\% &62.5\% &65.3\%  &92.3\% & \textbf{99.7}\%\\   
Luke   &1370 &624 & $53m^3$ & 54.65\% & 70.99\% & 46.0\%&83.3\% &39.3\% &50.5\% &71.5\% &\textbf{84.6}\%\\  
\hline
Floor5a &1001 &497 & $38m^3$ & 28.97\% & 38.43\% &49.5\%& 66.2\% &50.3\% &68.0\% &85.7\% &\textbf{89.9}\%\\   
Floor5b &1391 & 415&$79m^3$ & 56.87\% & 45.78\% & 56.4\%&71.1\% &58.5\% &83.2\% &92.3\% &\textbf{98.9}\%\\   
\hline
Gates362&2981 & 386&$29m^3$ & 49.48\% & 67.88\% &67.7\%&51.8\% &82.1\%&88.9\% &90.4\%&\textbf{96.7}\%\\    
Gates381&2949 &1053& $44m^3$ & 43.87\% & 62.77\% &54.6\%&52.3\%&51.4\% &73.9\% &82.3\% &\textbf{92.9}\%\\    
Lounge  &925 &327&$38m^3$ & 61.16\% & 58.72\% & 54.0\%&64.2\% &59.6\% &74.6\% &91.4\%&\textbf{94.8}\%\\    
Manolis &1613 &807 &$50m^3$ & 60.10\% & 72.86\% &65.1\%&76.0\%&68.5\% &87.2\% &93.9\% &\textbf{98.0}\%\\    
\hline
\textbf{Average} &--- &--- &--- & 53.7\%& 62.2\% &59.9\%&69.3\% &63.6\% &75.6\% & 87.1\% &\textbf{92.7}\%\\   
\hline
\end{tabular}
}
\end{center}
\caption{\textbf{Camera relocalization results for the indoor dataset.} We show the percentage of correct frames (within $5cm$ translational and $5^{\circ}$ angular error) of our method on 4 Scenes dataset against four state-of-the-art methods: ORB+PnP, SIFT+PnP, Random+SIFT \cite{LiliRandom}, and MNG \cite{valentin2016learning}. The best performance is highlighted.}
\label{table:rgbd_4scenes}
\end{table*}

\subsubsection*{\bf Dataset} The $4 \ Scenes$ dataset was introduced by Valentin \etal \cite{valentin2016learning} to push the boundaries of RGB-D and RGB camera relocalization. The recorded environments are significantly larger than the $7 \ Scenes$ dataset \cite{shotton2013scene} ($14-79 m^3$ versus $2-6m^3$). This large environment is more practical for the application of indoor robot localization. The scenes were captured by a Structure.io depth sensor with an iPad RGB camera. Both cameras have been calibrated and temporally synchronized. The RGB image sequences were recorded at a resolution of $1296 \times 968$ pixels, and the depth resolution is $640 \times 480$. We re-sampled the RGB images to the depth image resolution to align these images. The ground truth camera poses were from BundleFusion \cite{dai2016bundle}, which is a real-time globally consistent 3D reconstruction system. We use the training/testing dataset splitting from \cite{valentin2016learning}.

\subsubsection*{\bf Baselines and Error Metric} We use four state-of-the-art methods as our baselines: SIFT+PnP, ORB+PnP, Random+SIFT \cite{LiliRandom}, Multiscale Navigation Graph (MNG) \cite{valentin2016learning}. The SIFT+PnP and ORB+PnP are based on matching local features, and the results are from \cite{valentin2016learning}. These methods have to store a large database of descriptors. Random+SIFT \cite{LiliRandom} uses both random feature and sparse feature but in two separate steps. The sparse feature is used as a post-processing in this method. Our method is different from this method by simultaneously using both random feature and sparse features within the framework of regression forests. The multiscale navigation graph (MNG) method \cite{valentin2016learning} estimates the camera pose by maximizing the photo-consistency between the query frame and the synthesized image which is conditioned on the camera pose. At test time, the MNG method requires the trained model (retrieval forests) and the 3D model of the scene.

We report the percentage of test frames for which the estimated camera pose is essentially \singleQuote{correct}. A pose is correct when it is within $5 cm$ translational error and $5^o$ angular error of the ground truth. This accuracy is sufficient for camera relocation in restarting the camera tracking system and virtual reality \cite{shotton2013scene}.

\subsubsection*{\bf Image Features on RGB-D Images}
For the indoor dataset, we use random features and Walsh-Hadamard transform (WHT) features \cite{hel2005real}. Random features and WHT features do not require expensive feature detection so they can speed up the process. They also provide a sufficient number of robust features in texture-less areas, which is important for indoor camera relocalization.

In our method, random features are used for splitting decision trees in internal nodes, while WHT features are used to describe local patches in leaf nodes. 
The random features are based on pairwise pixel comparison as in \cite{shotton2013scene,valentin2015exploiting}:
\begin{equation}
f_{\mathbf{\phi}}(\mathbf{p}) = \mathtt{I} (\mathbf{p}, c_1) - \mathtt{I} (\mathbf{p} + \frac{\mathbf{\delta}}{\mathtt{D}(\mathbf{p})}, c_2)
\label{equ:feature}
\end{equation}
where $\mathbf{\delta}$ is a 2D offset and $\mathtt{I} (\mathbf{p}, c)$ indicates an RGB pixel lookup in channel $c$. The $\mathtt{D}(\mathbf{p})$ is a depth lookup. The $\mathbf{\phi}$ contains feature response parameters $\{\mathbf{\delta}, c_1, c_2 \}$.

The WHT features that we use contain 20 components for each RGB channel, thus our total feature dimension is 60 for all the three channels.

\subsubsection*{\bf Main Results and Analysis}
We present the main camera relocalization results in Table \ref{table:rgbd_4scenes}.  Besides our final method BTBRF, we also report the results of the unbalanced regression forest (UBRF) which does not use the sample-balanced objective, and the balanced regression forest (BRF) which uses both the sample-balanced objective and the spatial-variance objective in training but does not use backtracking in testing. 

In this dataset, sparse baselines do a reasonable job and the SIFT feature method is better than the ORB feature method. Our method using RGB-only images at test time achieves higher accuracy than SIFT+PnP, ORB+PnP, and Random+Sparse, and is less accurate than the MNG method. However, the MNG method needs an explicit 3D model to render synthetic images to refine the pose while our method does not need. Moreover, the MNG method needs a large number
of synthetic images (9 times of the original training images) for data augmentation while the present method does not need that. Our method using RGB-D images at test time considerably outperforms all the baselines in accuracy for camera relocalization. 

\begin{figure*}
	\begin{center}
		\includegraphics[width = \linewidth]{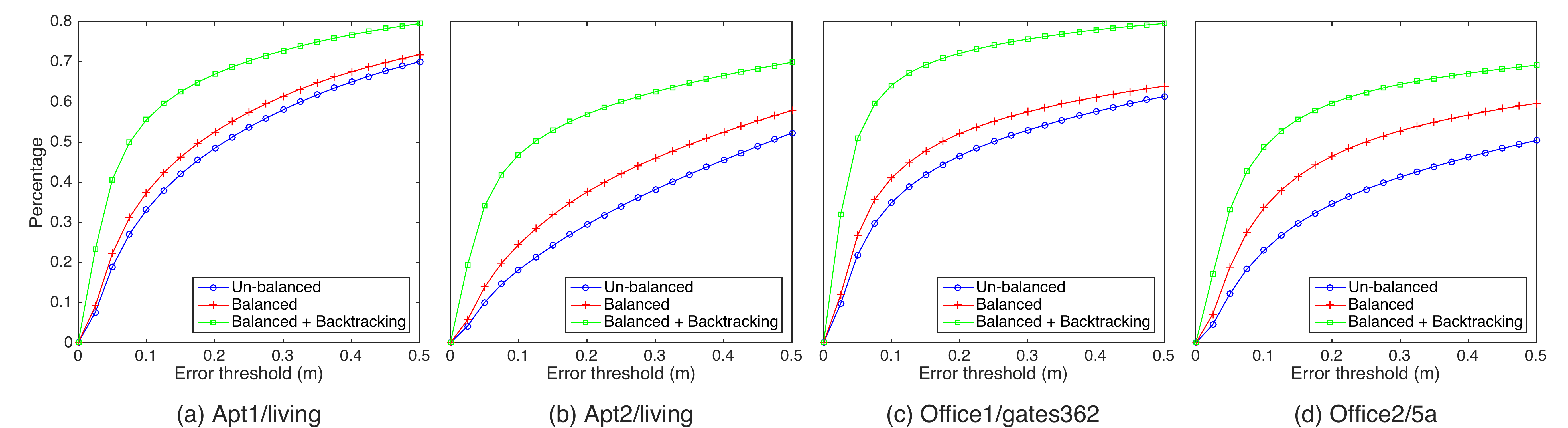}  
	\end{center}
	\vspace{-3mm}
	\caption{\textbf{Impact of the sample-balanced objective and backtracking on scene points prediction accuracy.} These figures show the accumulated percentage of predictions within a sequence of inlier thresholds. Please note this accuracy is not the final camera pose accuracy. The proposed method with the sample-balanced objective (red lines) consistently has a higher percentage of inliers compared with the unbalanced objective (blue lines). Backtracking (green lines) further improves prediction accuracy. The maximum number of backtracking leaves is 16 here.} 
	\label{fig:prediciton_error}
    \vspace{-1mm}
\end{figure*}

\begin{figure}
	\begin{center}
		\includegraphics[width = 0.8\linewidth]{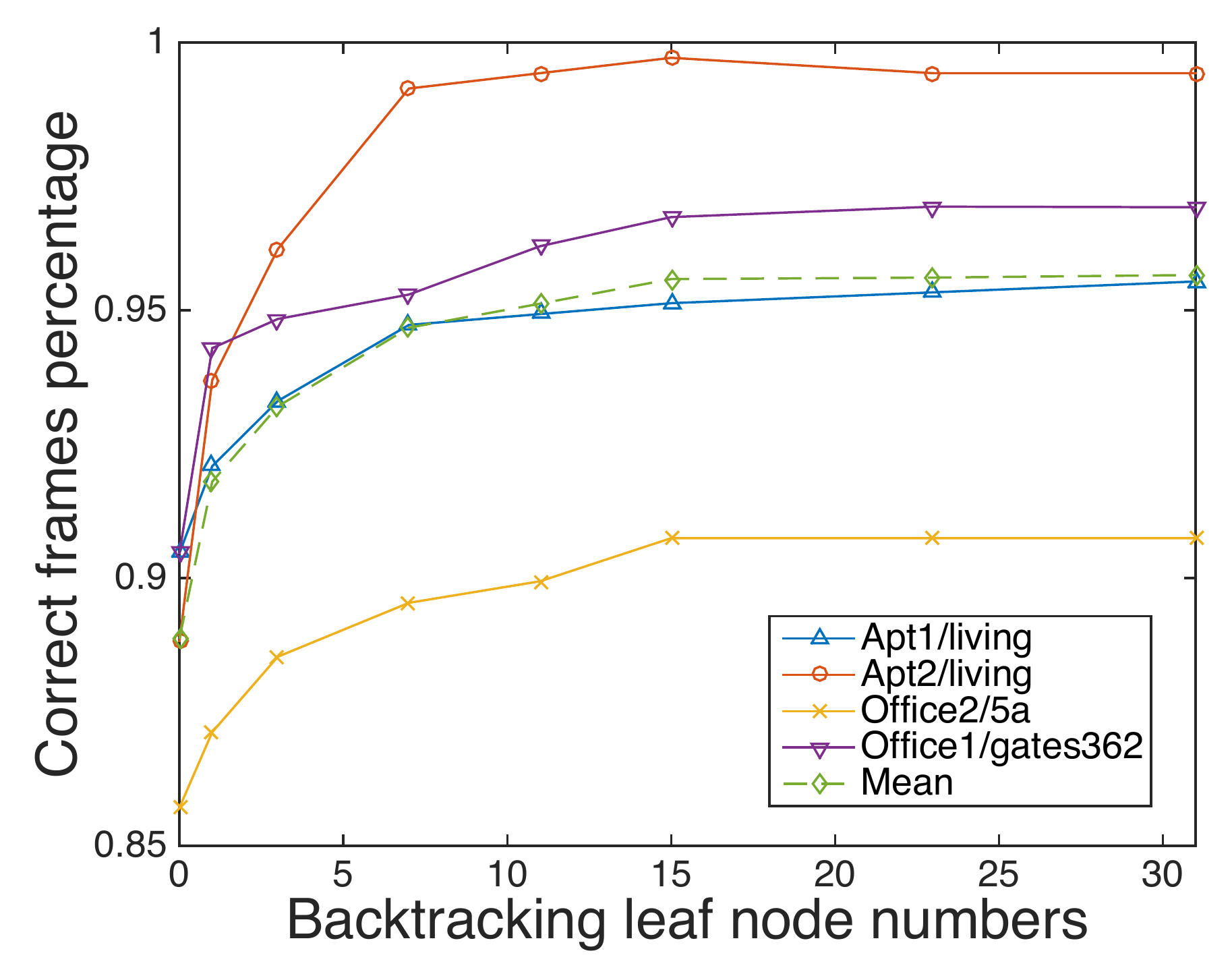}  
	\end{center}
	\vspace{-4mm}
	\caption{\textbf{Camera relocalization accuracy vs. backtracking leaf node numbers.} The camera relocalization performance increases with more backtracking leaf nodes though eventually levels out. } 
	\label{fig:backtracking}
    \vspace{-4mm}
\end{figure}

\begin{table*}
\begin{center}
\scalebox{1}{
\begin{tabular}{|l|l|cccccc|}
\hline
\textbf{Scene}  & \textbf{Spatial} & \textbf{Active Search} & \textbf{Active Search} &  \textbf{PoseNet }\cite{kendall2015posenet} & \textbf{Bayesian} &\textbf{CNN+}  & \textbf{BTBRF} \\
    &   \textbf{ Extent}  & \textbf{(VSfM)} \cite{sattler2016efficient} &\textbf{(COLMAP)}\cite{sattler2016efficient} & & \textbf{PoseNet}\cite{kendall2016modelling}& \textbf{LSTM}\cite{walch2016image}&\textbf{(Ours)} \\
\hline\hline
King's College & 140x40m &  0.67m, $0.52^{\circ}$&0.57m, $0.70^{\circ}$
 &1.92m, $5.40^{\circ}$&1.74m, $4.06^{\circ}$ &0.99m, $3.65^{\circ}$&\textbf{0.39}m, $\textbf{0.36}^{\circ}$ \\
Old Hospital & 50x40m & 1.29m, $0.79^{\circ}$&0.52m, $1.12^{\circ}$ &2.31m, $5.38^{\circ}$&2.57m, $5.14^{\circ}$& 1.51m, $4.29^{\circ}$& \textbf{0.30}m, $\textbf{0.41}^{\circ}$\\
Shop Facade & 35x25m & 0.17m, $0.53^{\circ}$& 0.12m, $0.41^{\circ}$ &1.46m, $8.08^{\circ}$&1.25m, $7.54^{\circ}$&1.18m, $7.44^{\circ}$&\textbf{0.15}m, $\textbf{0.31}^{\circ}$\\
St Mary's Church & 80x60m & 0.29m, $0.55^{\circ}$ &0.22m, $0.62^{\circ}$ &2.65m, $8.48^{\circ}$&2.11m, $8.38^{\circ}$&1.52m, $6.68^{\circ}$&\textbf{0.20}m, $\textbf{0.40}^{\circ}$\\
\hline
Average & &0.61m, $0.60^{\circ}$& 0.36m, $0.71^{\circ}$ &2.08m, $6.83^{\circ}$&1.92m, $6.28^{\circ}$&1.30m, $5.52^{\circ}$&\textbf{0.27}m, $\textbf{0.39}^{\circ}$\\
\hline
\end{tabular}
}
\end{center}
\caption{\textbf{Camera relocalization results for the outdoor Cambridge landmarks dataset}. We show median performance for our method against four state-of-the-art methods: Active Search with \cite{sattler2016efficient} with Visual SfM and with COLMAP \cite{schoenberger2016sfm} for reconstruction, PoseNet \cite{kendall2015posenet}, Bayesian PoseNet \cite{kendall2016modelling}, and CNN+LSTM \cite{walch2016image}. }
\label{table:median_cambridgeLandmarks}
\vspace{-2mm}
\end{table*}

To demonstrate that the improvement is indeed from the sample-balanced objective and backtracking, we compare the world coordinate prediction accuracy from our UBRF, BRF and BTBRF. Fig. \ref{fig:prediciton_error} shows the accumulated percentage of predictions within error thresholds for four sequences. For a particular threshold, the higher the percentage, the more accurate the prediction is. The figure clearly shows that our method with the sample-balanced objective and the backtracking technique is consistently better than the method without the sample-balanced objective and without backtracking.

We analyze the influences of the backtracking number $N_{max}$. Fig. \ref{fig:backtracking} shows the camera relocalization accuracy against different $N_{max}$. The accuracy is significantly improved within the first (about) 10 leaf nodes and saturates after that. Because the processing time linearly increases with the backtracking number, a small number of $N_{max}$ is preferred. 

\subsection{Outdoor Camera Relocalization}

\subsubsection*{\bf Dataset} The Cambridge landmarks dataset \cite{kendall2015posenet} is used to evaluate the developed method. It consists of data in a large scale outdoor urban environment. Scenes are recorded by a smart phone RGB camera at $1920 \times 1080$ resolution. The dataset also contains the structure-from-motion (SfM) models reconstructed with all images to get the ground truth camera pose. The dataset exhibits motion blur, moving pedestrians and vehicles, and occlusion, which pose great challenges for camera relocalization. 

\subsubsection*{\bf Baselines and Error Metric} Four state-of-the-art methods are used as the baselines: Active Search \cite{sattler2016efficient}, PoseNet \cite{kendall2015posenet}, Bayesian PoseNet \cite{kendall2016modelling}, and CNN+LSTM \cite{walch2016image}. Active Search employs a prioritized matching step that first considers features more likely to yield 2D-to-3D matches and then terminates the correspondence search once enough matches have been found. PoseNet trains a ConvNet as a pose regressor to estimate the 6-DOF camera pose from a single RGB image. Bayesian PoseNet improved the accuracy of PoseNet through an uncertainty framework by averaging Monte Carlo dropout samples from the posterior Bernoulli distribution of the Bayesian ConvNet's weights. CNN+LSTM \cite{walch2016image} uses a CNN to learn feature representations and LSTM units on the CNN output in spatial coordinates to capture the contextual information. The median translational error and rotational error are used here as in previous work \cite{kendall2015posenet, kendall2016modelling,walch2016image} for fair comparison.

\subsubsection*{\bf Image Features on RGB Images}
We use SIFT features \cite{lowe2004distinctive} as local feature descriptors for the outdoor camera relocalization. Because the BTBRF method has to store the mean vector of feature descriptors in leaf nodes, 64-dimension SIFT features are used to decrease the trained model size. In training, the 3D points of the SfM model are first projected to their corresponding images using ground truth camera poses. Then,  SIFT features are associated with their corresponding 3D world coordinates using the nearest neighbor (between SIFT feature 2D locations and projected 2D locations) within a threshold (1 pixel). In testing, SIFT features are detected from images. Their 3D locations are predicted by the regression forests. Dubious correspondences (feature distance $> 0.5$) are removed before the preemptive RANSAC.

\subsubsection*{\bf Main Results and Analysis}
Table \ref{table:median_cambridgeLandmarks} shows the median camera relocalization errors for the present method and the baseline methods. The results of PoseNet and Bayesian PoseNet are from the original papers. The results of active search and CNN+LSTM are from \cite{walch2016image}. The present method considerably outperforms all these baselines for all scenes in terms of median translational and rotational errors. It is about an order of magnitude improvement in accuracy compared with PoseNet, five times as accurate as CNN+LSTM and twice as accurate as Active Search using the same visual SfM \cite{wu2011visualsfm} reconstruction. Timings are not directly compared here as all these baselines are implemented on different high-end GPUs while our current implementation is on a single CPU core. 

\begin{figure*}
\centering
\begin{minipage}{0.41\linewidth}
\centering
\includegraphics[width=\linewidth]{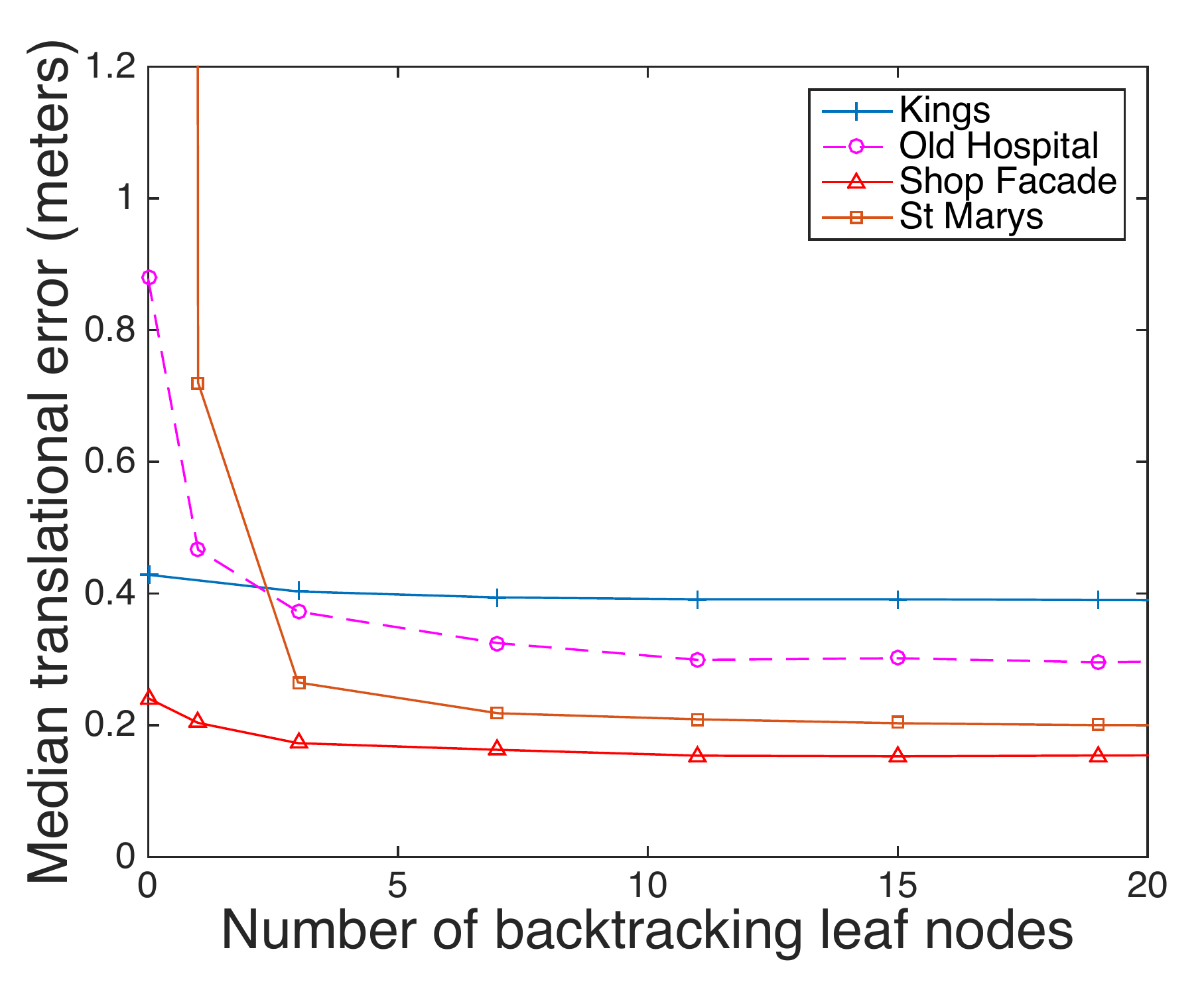}\\
\vspace{-1mm}
(a)
\end{minipage}
\begin{minipage}{0.40\linewidth}
\centering
\includegraphics[width=\linewidth]{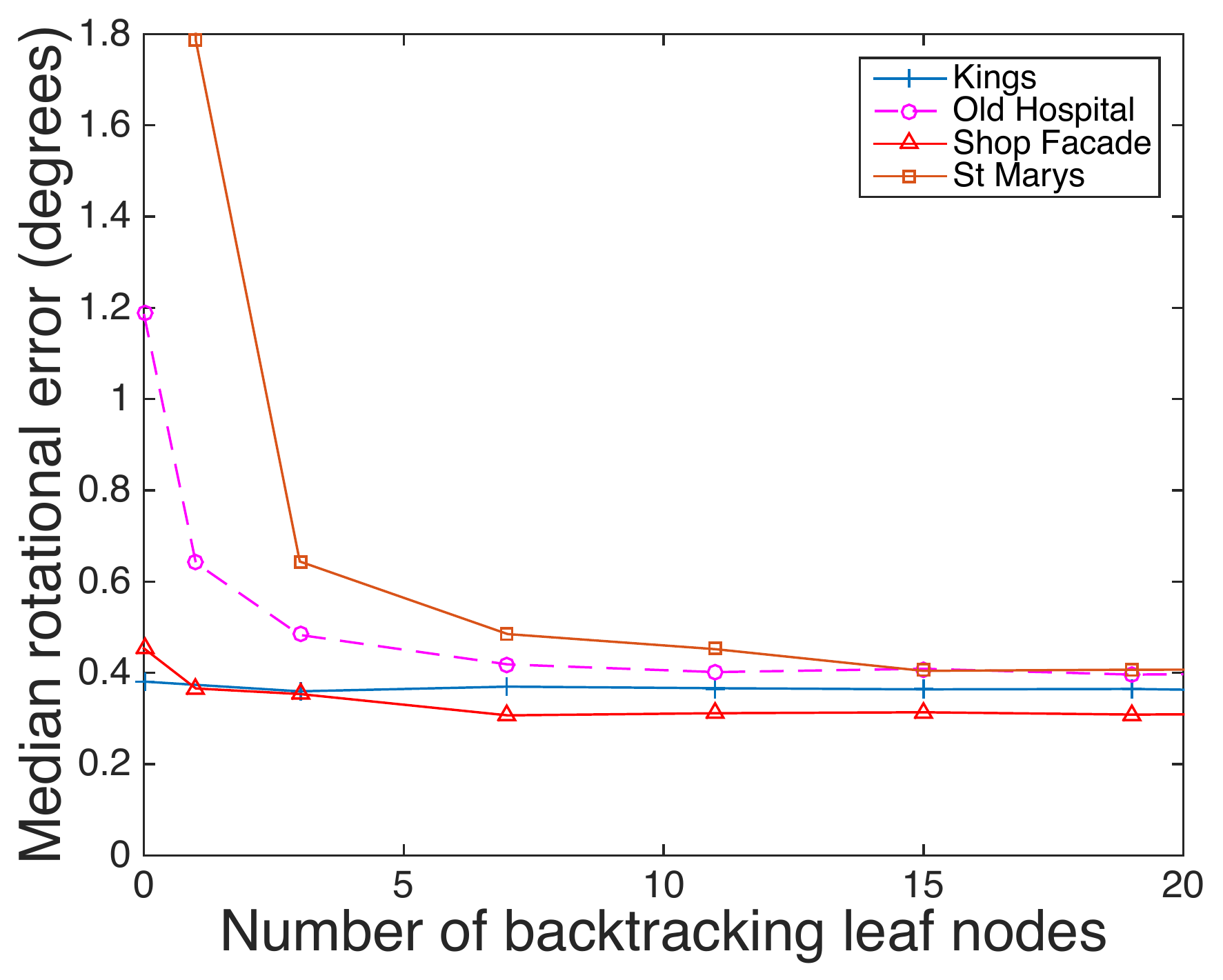} \\
\vspace{-1mm}
(b)  
\end{minipage}  
\vspace{-1mm}
\caption
  {\textbf{Effect of backtracking leaf node numbers on camera relocalization accuracy on Cambridge landmarks dataset. } (a) median translational error vs. backtracking leaf numbers (b) median rotational error vs. backtracking leaf numbers. Both median translational and rotational errors decrease with more backtracking leaf nodes though eventually level out. }
        \label{fig:BT_number_outdoor}
        \vspace{-3mm}
\end{figure*}
To gain some insight into how backtracking helps to improve the accuracy, Fig. \ref{fig:BT_number_outdoor} plots the camera relocalization accuracy against backtracking leaf node numbers. It shows that the camera relocalization errors significantly decrease within about 5 backtracking leaf nodes, which indicates that the relocalization accuracy can be improved by backtracking only a small number of leaf nodes.

\subsection{Qualitative Results}

\begin{figure*}
\centering
\begin{minipage}{0.4\linewidth}
\centering
\includegraphics[width=\linewidth]{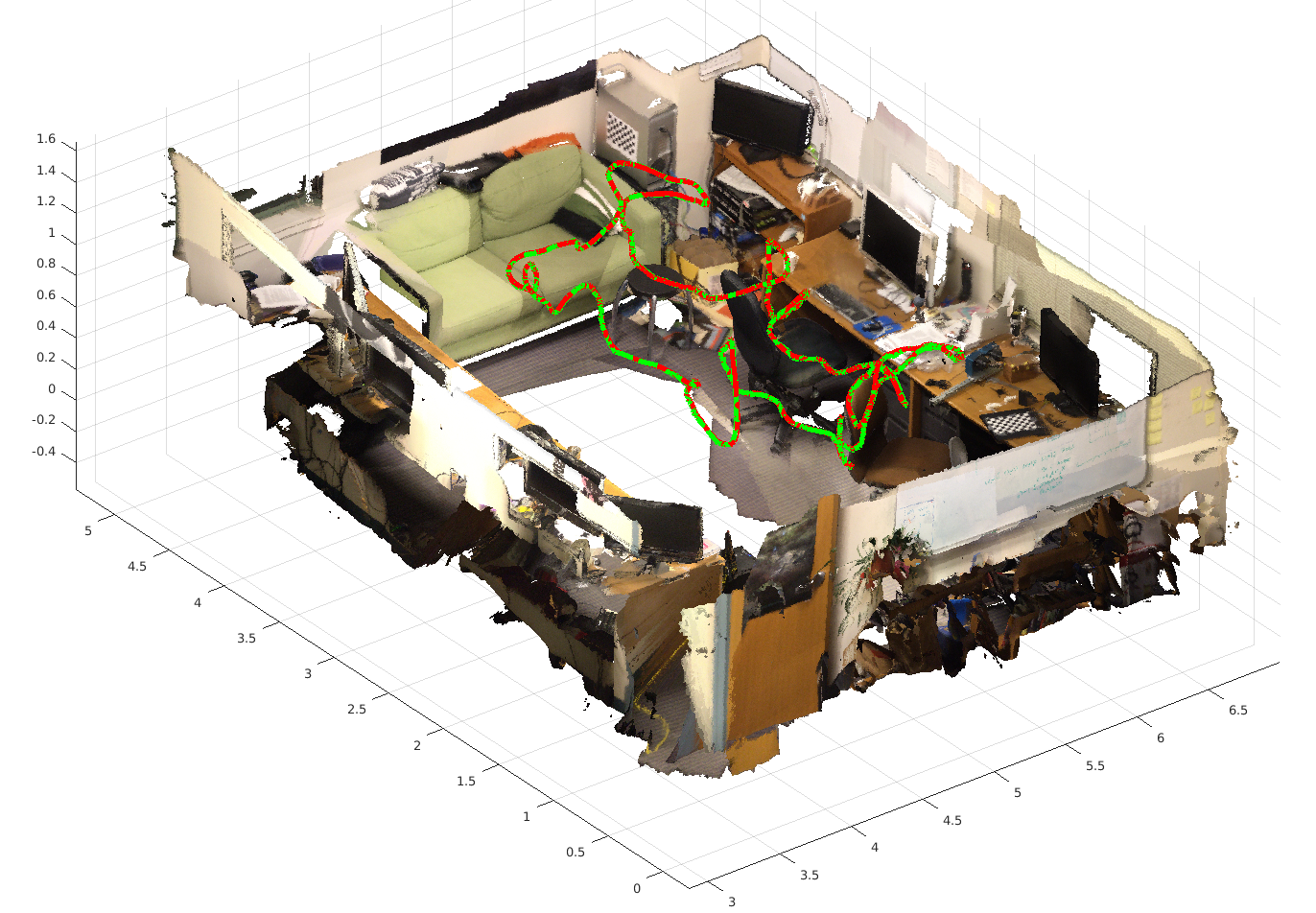}\\
\vspace{-1mm}
(a)
\end{minipage}
\begin{minipage}{0.4\linewidth}
\centering
\includegraphics[width=\linewidth]{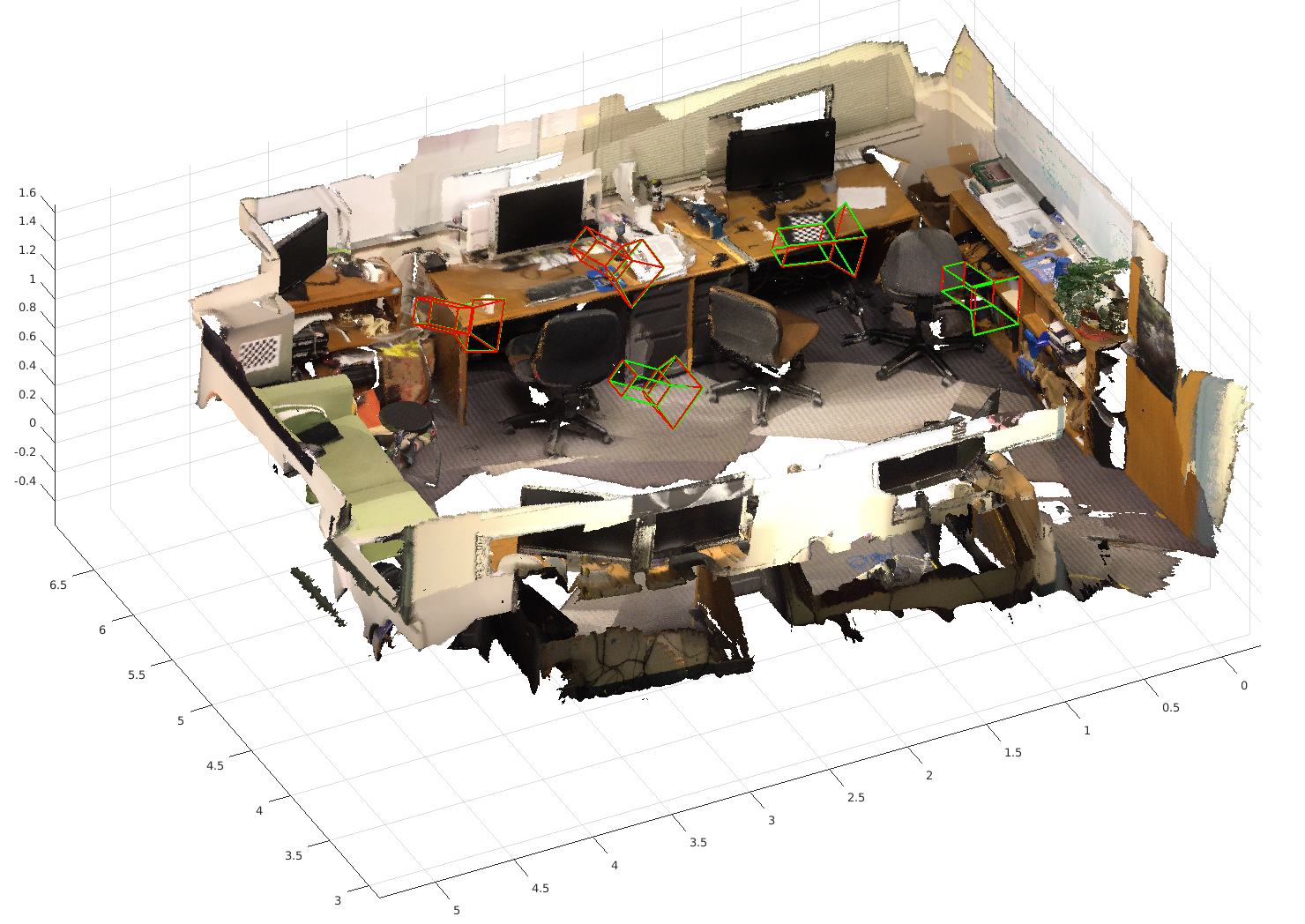} \\
\vspace{-1mm}
(b)  
\end{minipage}

\caption{\textbf{Qualitative results for indoor dataset (from office2/gates381).} Best viewed in color. The ground truth is in red and our estimated camera pose is in green. (a) camera trajectories. (b) several evenly sampled camera frusta are shown for visualization. Our method produces accurate camera locations and orientations. Note: the 3D model is only for visualization purposes and it is not used for our camera relocalization. }
\label{fig:qualitative_result_indoor}
\end{figure*}

\begin{figure*}
\centering
\includegraphics[width= 0.75\linewidth]{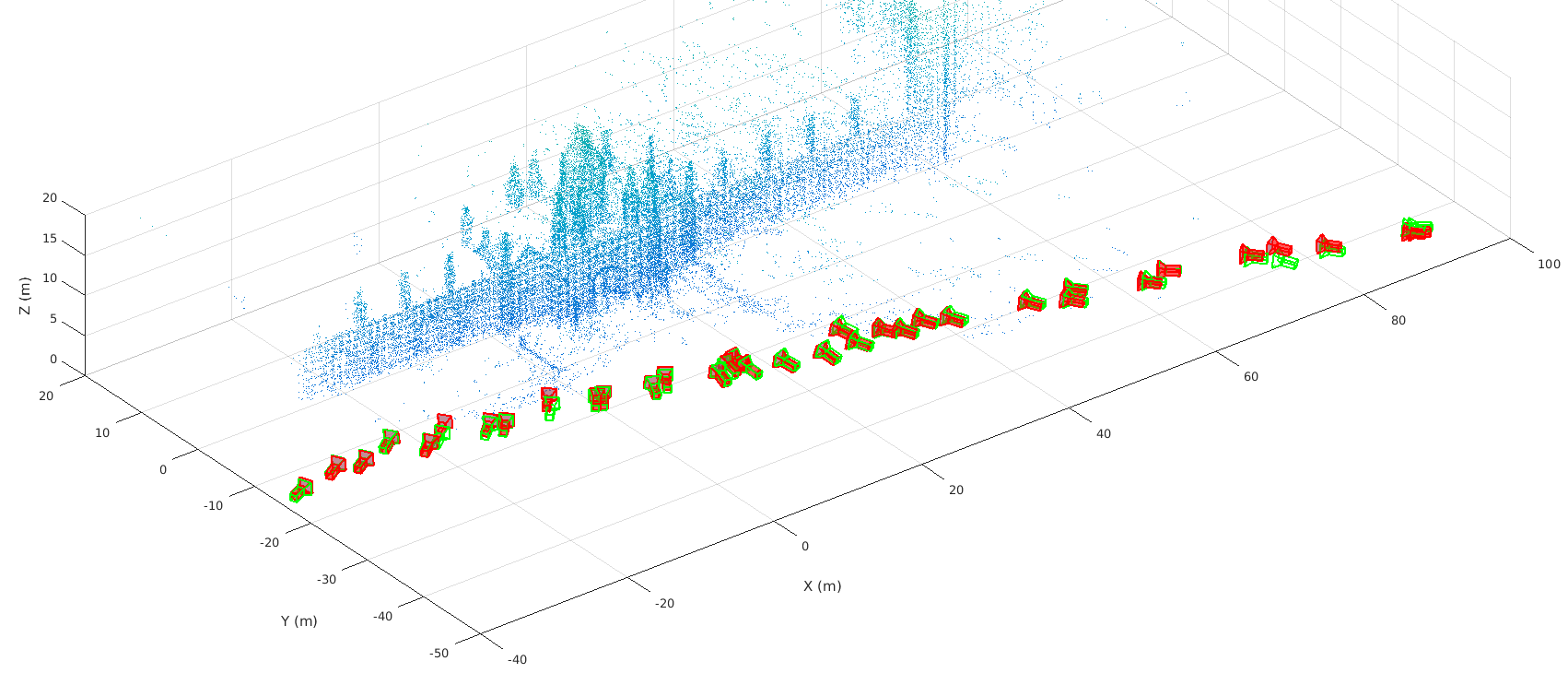}
\caption{\textbf{Qualitative results for the outdoor dataset Cambridge landmarks, King's College.} Best viewed in color. The camera poses are evenly sampled every ten frames for visualization. Camera frusta with hollow green frames show ground truth camera poses, while red ones with light opacity show our estimated camera poses. }
\label{fig:qualitive_results}
\vspace{-0.1in}
\end{figure*}

Fig. \ref{fig:qualitative_result_indoor} shows examples of qualitative results of our method in indoor scenes. In Fig. \ref{fig:qualitative_result_indoor} (a), the estimated camera trajectory and the ground truth trajectory are highly overlapped in the scene, illustrating the accuracy of predicted camera locations. In Fig. \ref{fig:qualitative_result_indoor} (b), several camera frusta show the accuracy of estimated camera orientation. These results show that the estimated camera poses are accurate in both location and orientation. Fig. \ref{fig:qualitive_results} shows the qualitative results for the King's College scene. The camera poses are uniformly sampled every ten frames to improve the visualization.

\subsection{Implementation Details}
The proposed method is implemented with C++ using OpenCV on an Intel 3GHz i7 CPU, 16GB memory Mac system. The node is available online {\footnote{\url{https://github.com/LiliMeng/btrf}}. The parameter settings for regression forest are: tree number $T=5$; 500 (indoor) and 300 (outdoor) training images per tree; 5,000 randomly sampled pixels per training image (indoor); the number of SIFT features per image varies from around 500 to 1,500 (outdoor); the maximum depth of tree is 25; the maximum backtracking leaves is 16. For the test, the present unoptimized implementation takes approximately 1 second (indoor) and 2 seconds (outdoor) on a single CPU core. Most of the time is on the backtracking and the camera pose optimization part. It should be noted that the current implementation is not optimized for speed and no GPU is used, which makes it possible to speed up with more engineering effort and GPU implementation \cite{sharp2008implementing}.

\subsection{Limitations}
At the moment, the proposed BTBRF method has two limitations that should be improved in the future. First, it has to store mean local patch descriptors in leaf nodes, which increases the trained model size.  Second, the time complexity of the backtracking is proportional to the value of $N_{max}$, making the prediction process slower than the conventional regression forests method. We would like to look for ways of adaptively setting the value of $N_{max}$ to avoid unnecessary backtracking.

\section{Conclusions}
In this paper, we proposed a sample-balanced objective and a backtracking scheme to reduce the effect of greedy splitting and improve the accuracy for camera relocalization. The proposed methods are applicable to regression forests both for RGB-D and RGB camera relocalization, showing state-of-the-art accuracy in publicly available datasets. Furthermore, we integrated local features in regression forests, advancing regression forests based methods to use RGB-only images for both training and test, and broadening the application to outdoor scenes while achieving the best accuracy against several strong state-of-the-art baselines. For future work, implementing on a parallel computation architecture will make the current work more efficient. Moreover, it will be interesting to integrate our camera relocalization to an autonomous robot navigation system \cite{autonomous_IROS2017} .   

\section{Acknowledgement}
The authors would like to thank The Institute for Computing, Information and Cognitive Systems (ICICS) Member Support Fund in University of British Columbia, Canada.







\bibliographystyle{IEEEtran}
\bibliography{iros2017}

\begin{thebibliography}{10}
\providecommand{\url}[1]{#1}
\csname url@samestyle\endcsname
\providecommand{\newblock}{\relax}
\providecommand{\bibinfo}[2]{#2}
\providecommand{\BIBentrySTDinterwordspacing}{\spaceskip=0pt\relax}
\providecommand{\BIBentryALTinterwordstretchfactor}{4}
\providecommand{\BIBentryALTinterwordspacing}{\spaceskip=\fontdimen2\font plus
\BIBentryALTinterwordstretchfactor\fontdimen3\font minus
  \fontdimen4\font\relax}
\providecommand{\BIBforeignlanguage}[2]{{%
\expandafter\ifx\csname l@#1\endcsname\relax
\typeout{** WARNING: IEEEtran.bst: No hyphenation pattern has been}%
\typeout{** loaded for the language `#1'. Using the pattern for}%
\typeout{** the default language instead.}%
\else
\language=\csname l@#1\endcsname
\fi
#2}}
\providecommand{\BIBdecl}{\relax}
\BIBdecl

\bibitem{shotton2013scene}
J.~Shotton, B.~Glocker, C.~Zach, S.~Izadi, A.~Criminisi, and A.~Fitzgibbon,
  ``Scene coordinate regression forests for camera relocalization in {RGB-D}
  images,'' in \emph{CVPR}, 2013.

\bibitem{guzman2014multi}
A.~Guzman-Rivera, P.~Kohli, B.~Glocker, J.~Shotton, T.~Sharp, A.~Fitzgibbon,
  and S.~Izadi, ``Multi-output learning for camera relocalization,'' in
  \emph{CVPR}, 2014.

\bibitem{valentin2015exploiting}
J.~Valentin, M.~Nie{\ss}ner, J.~Shotton, A.~Fitzgibbon, S.~Izadi, and P.~H.
  Torr, ``Exploiting uncertainty in regression forests for accurate camera
  relocalization,'' in \emph{CVPR}, 2015.

\bibitem{Brachmann_2016_CVPR}
E.~Brachmann, F.~Michel, A.~Krull, M.~Ying~Yang, S.~Gumhold, and c.~Rother,
  ``Uncertainty-driven {6D} pose estimation of objects and scenes from a single
  {RGB} image,'' in \emph{CVPR}, 2016.

\bibitem{LiliRandom}
L.~Meng, J.~Chen, F.~Tung, J.~J. Little, and C.~W. de~Silva, ``Exploiting
  random {RGB} and sparse features for camera pose estimation,'' in
  \emph{BMVC}, 2016.

\bibitem{kendall2015posenet}
A.~Kendall, M.~Grimes, and R.~Cipolla, ``{PoseNet}: A convolutional network for
  real-time {6-DOF} camera relocalization,'' in \emph{ICCV}, 2015.

\bibitem{kendall2016modelling}
A.~Kendall and R.~Cipolla, ``Modelling uncertainty in deep learning for camera
  relocalization,'' in \emph{ICRA}, 2016.

\bibitem{walch2016image}
F.~Walch, C.~Hazirbas, L.~Leal-Taix{\'e}, T.~Sattler, S.~Hilsenbeck, and
  D.~Cremers, ``Image-based localization with spatial {L}{S}{T}{M}s,''
  \emph{arXiv preprint arXiv:1611.07890}, 2016.

\bibitem{sattler2016efficient}
T.~Sattler, B.~Leibe, and L.~Kobbelt, ``Efficient \& effective prioritized
  matching for large-scale image-based localization,'' \emph{IEEE Trans on
  PAMI}, 2016.

\bibitem{nister2006scalable}
D.~Nister and H.~Stewenius, ``Scalable recognition with a vocabulary tree,'' in
  \emph{CVPR}, 2006.

\bibitem{torii201524}
A.~Torii, R.~Arandjelovic, J.~Sivic, M.~Okutomi, and T.~Pajdla, ``24/7 place
  recognition by view synthesis,'' in \emph{CVPR}, 2015.

\bibitem{klein2007parallel}
G.~Klein and D.~Murray, ``Parallel tracking and mapping for small {AR}
  workspaces,'' in \emph{Mixed and Augmented Reality.}, 2007.

\bibitem{glocker2015real}
B.~Glocker, J.~Shotton, A.~Criminisi, and S.~Izadi, ``Real-time {RGB-D} camera
  relocalization via randomized ferns for keyframe encoding,''
  \emph{Visualization and Computer Graphics, IEEE Trans. on}, 2015.

\bibitem{whelan2015elasticfusion}
T.~Whelan, S.~Leutenegger, R.~F. Salas-Moreno, B.~Glocker, and A.~J. Davison,
  ``Elasticfusion: Dense {SLAM} without a pose graph,'' \emph{RSS}, 2015.

\bibitem{se2005vision}
S.~Se, D.~G. Lowe, and J.~J. Little, ``Vision-based global localization and
  mapping for mobile robots,'' \emph{Robotics, IEEE Trans. on}, 2005.

\bibitem{cummins2011appearance}
M.~Cummins and P.~Newman, ``Appearance-only {S}{L}{A}{M} at large scale with
  {F}{A}{B}-{M}{A}{P} 2.0,'' \emph{IJRR}, 2011.

\bibitem{chen2017where}
J.~Chen and J.~J. Little, ``Where should cameras look at soccer games:
  Improving smoothness using the overlapped hidden markov model,''
  \emph{Computer Vision and Image Understanding}, vol. 159, pp. 59--73, 2017.

\bibitem{chen2015mimicking}
J.~Chen and P.~Carr, ``Mimicking human camera operators,'' in \emph{WACV},
  2015.

\bibitem{rubio2015efficient}
A.~Rubio, M.~Villamizar, L.~Ferraz, A.~Penate-Sanchez, A.~Ramisa,
  E.~Simo-Serra, A.~Sanfeliu, and F.~Moreno-Noguer, ``Efficient monocular pose
  estimation for complex 3d models,'' in \emph{ICRA}, 2015.

\bibitem{gao2003complete}
X.-S. Gao, X.-R. Hou, J.~Tang, and H.-F. Cheng, ``Complete solution
  classification for the perspective-three-point problem,'' \emph{PAMI, IEEE
  Trans. on}, 2003.

\bibitem{fischler1981random}
M.~A. Fischler and R.~C. Bolles, ``Random sample consensus: a paradigm for
  model fitting with applications to image analysis and automated
  cartography,'' \emph{Communications of the ACM}, 1981.

\bibitem{lowe2004distinctive}
D.~G. Lowe, ``Distinctive image features from scale-invariant keypoints,''
  \emph{IJCV}, 2004.

\bibitem{schindler2007city}
G.~Schindler, M.~Brown, and R.~Szeliski, ``City-scale location recognition,''
  in \emph{CVPR}, 2007.

\bibitem{lepetit2006keypoint}
V.~Lepetit and P.~Fua, ``Keypoint recognition using randomized trees,''
  \emph{IEEE trans on PAMI}, 2006.

\bibitem{gee20126d}
A.~P. Gee and W.~W. Mayol-Cuevas, ``{6D} relocalisation for {RGBD} cameras
  using synthetic view regression.'' in \emph{BMVC}, 2012.

\bibitem{engel2014lsd}
J.~Engel, T.~Sch{\"o}ps, and D.~Cremers, ``{LSD-SLAM}: Large-scale direct
  monocular {SLAM},'' in \emph{ECCV}, 2014.

\bibitem{valentin2016learning}
J.~Valentin, A.~Dai, M.~Nie{\ss}ner, P.~Kohli, P.~Torr, S.~Izadi, and
  C.~Keskin, ``Learning to navigate the energy landscape,'' in \emph{3DV},
  2016.

\bibitem{cavallari2017fly}
T.~Cavallari, S.~Golodetz, N.~A. Lord, J.~Valentin, L.~Di~Stefano, and P.~H.
  Torr, ``On-the-fly adaptation of regression forests for online camera
  relocalisation,'' in \emph{CVPR}, 2017.

\bibitem{redmon2016you}
J.~Redmon, S.~Divvala, R.~Girshick, and A.~Farhadi, ``You only look once:
  Unified, real-time object detection,'' in \emph{CVPR}, 2016.

\bibitem{kendall2017geometric}
A.~Kendall and R.~Cipolla, ``Geometric loss functions for camera pose
  regression with deep learning,'' in \emph{CVPR}, 2017.

\bibitem{breiman2001random}
L.~Breiman, ``Random forests,'' \emph{Machine learning}, 2001.

\bibitem{hel2005real}
Y.~Hel-Or and H.~Hel-Or, ``Real-time pattern matching using projection
  kernels,'' \emph{PAMI, IEEE Trans. on}, 2005.

\bibitem{kabsch1976solution}
W.~Kabsch, ``A solution for the best rotation to relate two sets of vectors,''
  \emph{Acta Crystallographica Section A: Crystal Physics, Diffraction,
  Theoretical and General Crystallography}, 1976.

\bibitem{nister2005preemptive}
D.~Nist{\'e}r, ``Preemptive {R}{A}{N}{S}{A}{C} for live structure and motion
  estimation,'' \emph{Machine Vision and Applications}, 2005.

\bibitem{dai2016bundle}
A.~Dai, M.~Nie{\ss}ner, M.~Zoll{\"o}fer, S.~Izadi, and C.~Theobalt,
  ``Bundlefusion: Real-time globally consistent {3D} reconstruction using
  on-the-fly surface re-integration,'' \emph{ACM Trans. on Graphics}, 2017.

\bibitem{schoenberger2016sfm}
J.~L. Sch\"{o}nberger and J.-M. Frahm, ``Structure-from-motion revisited,'' in
  \emph{CVPR}, 2016.

\bibitem{wu2011visualsfm}
C.~Wu \emph{et~al.}, ``Visual{SFM}: A visual structure from motion system,''
  2011.

\bibitem{sharp2008implementing}
T.~Sharp, ``Implementing decision trees and forests on a {G}{P}{U},'' in
  \emph{ECCV}, 2008.

\bibitem{autonomous_IROS2017}
C.~Wang, L.~Meng, S.~She, I.~M. Mitchell, T.~Li, F.~Tung, W.~Wan, M.~Q.~H.
  Meng, and C.~de~Silva, ``Autonomous mobile robot navigation in uneven and
  unstructured indoor environments,'' in \emph{IROS}, 2017.

\end{thebibliography}

\end{document}